\documentclass{article}

\usepackage{arxiv}

\usepackage[utf8]{inputenc} % allow utf-8 input
\usepackage[T1]{fontenc}    % use 8-bit T1 fonts
\usepackage{hyperref}       % hyperlinks
\usepackage{url}            % simple URL typesetting
\usepackage{booktabs}       % professional-quality tables
\usepackage{amsfonts}       % blackboard math symbols
\usepackage{nicefrac}       % compact symbols for 1/2, etc.
\usepackage{microtype}      % microtypography
\usepackage{lipsum}
\usepackage{xcolor}
\usepackage{listings}
\usepackage{bm}
\usepackage{graphicx}
\usepackage{rotating}
\usepackage{multirow}
\usepackage[linesnumbered,vlined]{algorithm2e}

\definecolor{codegreen}{rgb}{0,0.5,0}
\definecolor{codepurple}{rgb}{0.88,0,0.32}
\definecolor{backcolour}{rgb}{0.97,0.97,0.97}
 
\lstdefinestyle{coding}{
    backgroundcolor=\color{backcolour},   
    commentstyle=\color{codegreen},
    keywordstyle=\color{codepurple},
    basicstyle=\ttfamily,
    breakatwhitespace=false,         
    breaklines=true,                 
    captionpos=b,                    
    keepspaces=true,                 
    numbers=none,                    
    numbersep=5pt,                  
    showspaces=false,                
    showstringspaces=false,
    showtabs=false,                  
    tabsize=4
}
 
\title{Speeding Up OPFython with Numba}

\author{
  Gustavo H. de Rosa, João P. Papa \\
  Department of Computing \\
  São Paulo State University \\
  Bauru, São Paulo - Brazil \\
  \texttt{gustavo.rosa@unesp.br, joao.papa@unesp.br} \\
}
\begin{document}

% Making title
\maketitle

% Abstract
\begin{abstract}
A graph-inspired classifier, known as Optimum-Path Forest (OPF), has proven to be a state-of-the-art algorithm comparable to Logistic Regressors, Support Vector Machines in a wide variety of tasks. Recently, its Python-based version, denoted as OPFython, has been proposed to provide a more friendly framework and a faster prototyping environment. Nevertheless, Python-based algorithms are slower than their counterpart C-based algorithms, impacting their performance when confronted with large amounts of data. Therefore, this paper proposed a simple yet highly efficient speed up using the Numba package, which accelerates Numpy-based calculations and attempts to increase the algorithm's overall performance. Experimental results showed that the proposed approach achieved better results than the na\"ive Python-based OPF and speeded up its distance measurement calculation.
\end{abstract}

% Keywords

% Including external sections
\section{Introduction}
\label{s.intro}

Artificial Intelligence (AI) has provided a unique way to autonomously solve tasks in real-world scenarios, such as image classification, object recognition, medical analysis, text generation, among others~\cite{Russell:16}. Furthermore, it removed some burden from humans, which had to perform repetitive tasks or even define non-trivial decisions~\cite{Acemoglu:18}. Another important area, denoted as Machine Learning (ML), has received a great deal of attention due to its ability in modeling algorithms to accomplish suck tasks. Essentially, ML is responsible for fostering and implementing autonomous techniques, such as classifiers and neural networks capable of solving AI-based tasks~\cite{Bishop:06}.

Papa et al.~\cite{Papa:09,Papa:12} had proposed a state-of-the-art graph-based classifier known as Optimum-Path Forest (OPF), which aims to compose forests from optimal trees and conquer new nodes according to their corresponding patterns. Furthermore, the authors also have proposed a C-based package, denoted as LibOPF~\cite{LibOPF:15}, which comprehends all OPF's data structures and functional implementations. Recently, Rosa et al.~\cite{Rosa:20} have proposed a Python-based package, known as OPFython, aiming to provide a more friendly implementation and a faster prototyping workspace. Nevertheless, the OPF algorithm heavily depends on a distance measure repeatedly calculated between the nodes in the graph. Additionally, such calculations are highly cumbersome in high-level languages, such as Python. Thus, the main problem of the OPFython package is the amount of time it takes to be trained and tested when collated with large amounts of data.

This work addresses such a problem by using an additional Python-based package, denoted as Numba~\cite{Lam:15}, which aims at speeding up the execution of Numpy-based operations by performing low-level implementations. Additionally, such addition is straightforward to be implemented and showed that it is possible to improve the distance measure calculation in Python-based environments. Therefore, the main contributions of this technical report are two-fold: (i) to introduce Numba-based operations in OPFython, and (ii) to speed up OPF-based training and prediction in Python environments. The remainder of this work is organized as follows. First, Section~\ref{s.theory} presents a theoretical background regarding OPFython, as well as its distance measures. Section~\ref{s.methodology} introduces the proposed approach, employed datasets, and experimental setups, while Section~\ref{s.experiments} presents the experimental results. Finally, Section~\ref{s.conclusion} states the conclusions and future works.
\section{Theoretical Foundation}
\label{s.theory}

This section presents a brief overview of OPFython's architecture, as well as its available distance measures.

\subsection{OPFython}
\label{ss.opfython}

\begin{figure}[!ht]
\centering
\includegraphics[scale=0.6]{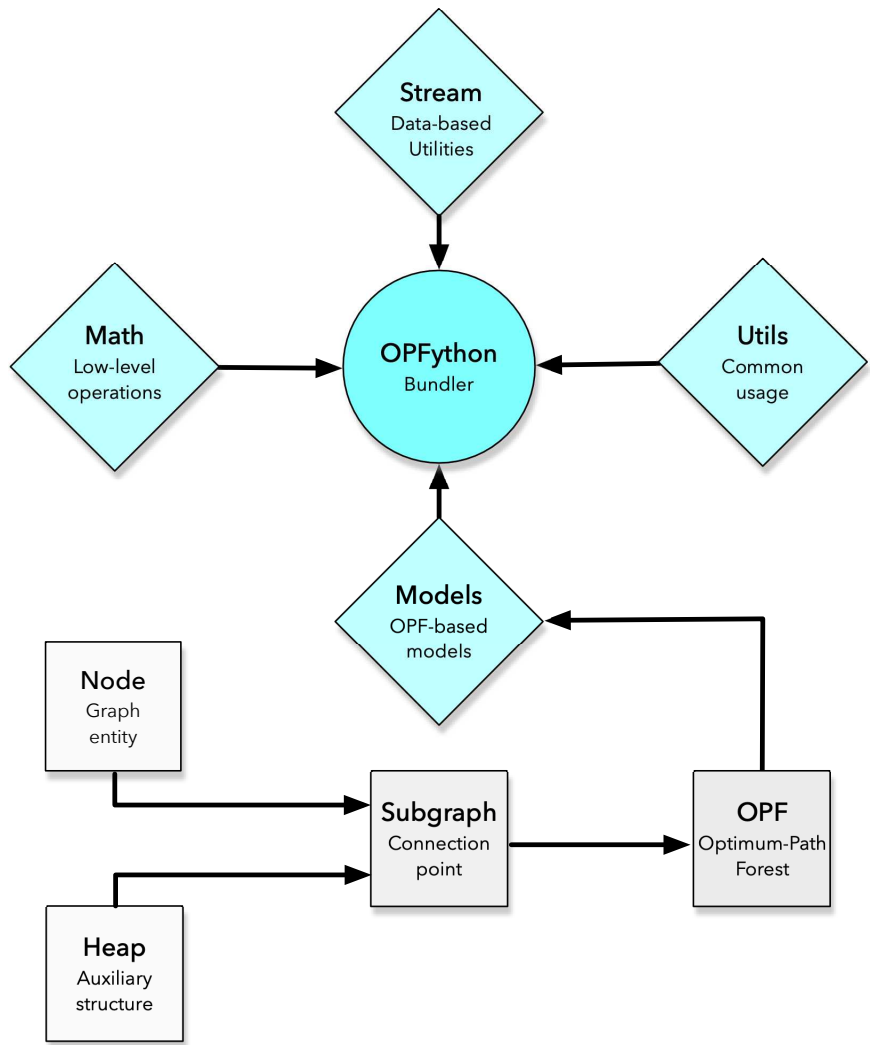}
\caption{Flowchart of OPFython's architecture.}
\label{f.flowchart}    
\end{figure}

OPFython is composed of several packages, each one responsible for appropriate classes and methods. Figure~\ref{f.flowchart} depicts a flowchart of OPFython's architecture, while the subsequent sections present each package.

\vspace*{2cm}

\clearpage

\subsubsection{Core}
\label{sss.core}

The `core` package implements all OPFython's base classes. It assists as a building foundation for developing more appropriate structures that may be required when creating an Optimum-Path Forest classifier. Four modules compose the core package, as follows:

\begin{itemize}

\item \textbf{Heap:} Assists OPF in stacking nodes' according to their costs and unstacking into the subgraph;

\item \textbf{Node:} Stores valuable information of a sample, such as its features, label, and other information that OPF might need;

\item \textbf{OPF:} Implements basic fitting and predicting methods, as well as assists in saving and loading pre-trained models;

\item \textbf{Subgraph:} Graph-based classifiers use subgraphs to build optimum-path costs and find the prototype nodes.

\end{itemize}

\subsubsection{Math}
\label{sss.math}

OPFython offers a `math` package that holds low-level mathematical implementations. Naturally, some constantly used functions are implemented in this package, as follows:

\begin{itemize}

\item \textbf{Distance:} Offers a variety of distance measures to calculate the cost between nodes;

\item \textbf{General:} Functions without categories;

\item \textbf{Random:} Random-based samplers.

\end{itemize}

\subsubsection{Models}
\label{sss.models}

Optimum-Path Forest classifiers can be implemented in various tasks, such as supervised, unsupervised, semi-supervised, among others. Thus, the `models` package provides classes and methods that compose these high-level strategies. Currently, there are four types of classifiers, as follows:

\begin{itemize}

\item \textbf{KNNSupervisedOPF~\cite{PapaKNN:09}:} Supervised Optimum-Path Forest with a KNN-based subgraph;

\item \textbf{SemiSupervisedOPF~\cite{Amorim:14}:} Semi-supervised Optimum-Path Forest;

\item \textbf{SupervisedOPF~\cite{Papa:09}:} Supervised Optimum-Path Forest;

\item \textbf{UnsupervisedOPF~\cite{Rocha:09}:} Unsupervised Optimum-Path Forest.

\end{itemize}

\subsubsection{Stream}
\label{sss.stream}

The `stream` is responsible for loading, parsing, and splitting data, being composed of the following modules:

\begin{itemize}

\item \textbf{Loader:} Assists users in pre-loading datasets in .txt, .csv and .json formats;

\item \textbf{Parser:} Parses the pre-loaded arrays into samples and labels;

\item \textbf{Splitter:} Splits loaded and parsed datasets into new sets.

\end{itemize}

\subsubsection{Subgraphs}
\label{sss.subgraphs}

One can observe that distinct classifiers might need distinct subgraphs. Therefore, the `subgraphs` package implements some additional classes, as follows:

\begin{itemize}

\item \textbf{KNNSubgraph:} KNN-based subgraph that works with KNN-based OPF.
\end{itemize}

\subsubsection{Utils}
\label{sss.utils}

The `utils` package implements standard functions shared over the library, as follows:

\begin{itemize}

\item \textbf{Constants:} Fixed values that do not alter throughout the code;

\item \textbf{Converter:} Converts .opf files into .txt, .csv, and .json files;

\item \textbf{Decorator:} Common functionalities wrappers;

\item \textbf{Exception:} Implements common errors and exceptions;

\item \textbf{Logging:} Logs crucial information regarding classes and functions.

\end{itemize}

\subsection{Distance Measures}
\label{ss.dist}

Distances are mathematical formulations used to define closeness or farness between entities and are usually calculated between two vectors $x$ and $y$, where $d(x,y)$ is a function that defines the distance between both vectors a non-negative real number. The depicted distances follow the taxonomy proposed by Alfeilat et al.~\cite{Abu:19}, being divided into eight categories as follows:

\begin{itemize}

\item $\mathbf{L_p}$\textbf{:} Chebyshev ($D_1$), Chi-Squared ($D_2$), Euclidean ($D_3$), Gaussian ($D_4$), Log-Euclidean ($D_5$), and Manhattan ($D_6$);

\item $\mathbf{L_1}$\textbf{:} Bray-Curtis ($D_7$), Canberra ($D_8$), Gower ($D_9$), Kulczynski ($D_{10}$), Lorentzian ($D_{11}$), Non-Intersection ($D_{12}$), and Soergel ($D_{13}$);

\item \textbf{Inner product:} Chord ($D_{14}$), Cosine ($D_{15}$), Dice ($D_{16}$), and Jaccard ($D_{17}$);

\item \textbf{Squared chord:} Bhattacharyya ($D_{18}$), Hellinger ($D_{19}$), Matusita ($D_{20}$), and Squared Chord ($D_{21}$);

\item \textbf{Squared $\mathbf{L_2}$:} Additive Symmetric $\mathcal{X}^2$ ($D_{22}$), Average Euclidean ($D_{23}$), Clark ($D_{24}$), Divergence ($D_{25}$), Log-Squared Euclidean ($D_{26}$), Mean Censored Euclidean ($D_{27}$), Neyman $\mathcal{X}^2$ ($D_{28}$), Pearson $\mathcal{X}^2$ ($D_{29}$), Sangvi $\mathcal{X}^2$ ($D_{30}$), Squared $\mathcal{X}^2$ ($D_{31}$), and Squared Euclidean ($D_{32}$);

\item \textbf{Shannon entropy:} Jeffreys ($D_{33}$), Jensen ($D_{34}$), Jensen-Shannon ($D_{35}$), K-Divergence ($D_{36}$), Kullback-Leibler ($D_{37}$), and Topsoe ($D_{38}$);

\item \textbf{Vicissitude:} Max Symmetric $\mathcal{X}^2$ ($D_{39}$), Min Symmetric $\mathcal{X}^2$ ($D_{40}$), Vicis Symmetric 1 ($D_{41}$), Vicis Symmetric 2 ($D_{42}$), Vicis Symmetric 3 ($D_{43}$), and Vicis-Wave Hedges ($D_{44}$);

\item \textbf{Other:} Hamming ($D_{45}$), Hassanat ($D_{46}$), and Statistic ($D_{47}$).
	
\end{itemize}
\section{Methodology}
\label{s.methodology}

This section presents the proposed approach, as well as the employed datasets and experimental setup.

\subsection{Proposed Approach}
\label{ss.proposed_approach}

Optimum-Path Forests are graph-based classifiers and rely on distance functions to estimate the cost of arcs. The first step is to measure the distance of every pair of nodes and compare it to the graph, where the best paths (minimum costs) will be defined as prototypes. Further, for every new node inserted in the graph, it will have its distance calculated and compared to the previously defined prototypes, being conquered by the prototype that offers the best path.

Let $\mathcal{D}$ be the set of available distances, $\mathcal{Z}_1$ and $\mathcal{Z}_2$ be the training and testing sets of a particular dataset, respectively, and $O$ be the OPF classifier. After training and evaluating the classifiers using different distance measures, the training time will be assessed and compared amongst themselves. Algorithm~\ref{a.proposed} describes the pseudo-code of the proposed approach.

\begin{algorithm}[!ht]
\KwIn{Training set $\mathcal{Z}_1$, 	testing set $\mathcal{Z}_2$, set of distances $\mathcal{D}$ and OPF classifier $O$.}
\KwOut{Training time $t$ and prediction time $p$.}
\For{$d \in \mathcal{D}$}{
	$O, t \leftarrow$ train over $\mathcal{Z}_1$ with distance $d$\;
	$p \leftarrow$ evaluate $O$ over $\mathcal{Z}_2$\;
}
\caption{Proposed approach pseudo-code.}
\label{a.proposed} 
\end{algorithm}

\subsection{Datasets}
\label{ss.datasets}

The proposed approach aims at assessing both Optimum-Path Forest and Numba-based Optimum-Path Forest training times in supervised classification tasks, which is currently implemented in the OPFython~\cite{Rosa:20} library. Table~\ref{t.datasets} describes an overview concerning the $22$ employed datasets, their task type, and their number of samples and features. Note that such a large amount of datasets attempt to overcome the diversity of problems, i.e., low/high number of features, small/large amounts of samples, and distinct domains.

\begin{table}[!ht]
	\renewcommand{\arraystretch}{1.75}
	\setlength{\tabcolsep}{10pt}
	\centering
	\caption{Employed datasets used in the computations.}
    \label{t.datasets}
	\begin{tabular}{cccc}
		\toprule
		\textbf{Dataset} & \textbf{Task} & \textbf{Samples} & \textbf{Features}
		\\ \midrule
		Arcene & Mass Spectrometry & $200$ & $10,000$
		\\ \midrule
		BASEHOCK & Text & $1,993$ & $4,862$
		\\ \midrule
		Caltech101 & Image Silhouettes & $8,671$ & $784$
		\\ \midrule
		COIL20 & Face Image & $1,540$ & $1,024$
		\\ \midrule                 
		Isolet & Spoken Letter Recognition & $1,560$ & $617$
		\\ \midrule
		Lung & Biological & $203$ & $3,312$
		\\ \midrule
		Madelon & Artificial & $2,600$ & $500$
		\\ \midrule
		MPEG7 & Image Silhouettes & $1,400$ & $1,024$
		\\ \midrule
		MPEG7-BAS & Image Descriptor & $1,400$ & $180$
		\\ \midrule
		MPEG7-Fourier & Image Descriptor & $1,400$ & $126$
		\\ \midrule
		Mushrooms & Biological & $8,124$ & $112$
		\\ \midrule
		NTL-Commercial & Energy Theft & $4,952$ & $8$
		\\ \midrule
		NTL-Industrial & Energy Theft & $3,182$ & $8$
		\\ \midrule
		ORL & Face Image & $400$ & $1,024$
		\\ \midrule
		PCMAC & Text & $1,943$ & $3,289$
		\\ \midrule
		Phishing & Network Security & $11,055$ & $68$
		\\ \midrule
		Segment & Image Segmentation & $2,310$ & $19$
		\\ \midrule
		Semeion & Handwritten Digits & $1,593$ & $256$
		\\ \midrule
		Sonar & Signal & $208$ & $60$
		\\ \midrule
		Spambase & Network Security & $4,601$ & $48$
		\\ \midrule
		Vehicle & Image Silhouettes & $846$ & $18$
		\\ \midrule
		Wine & Chemical & $178$ & $13$
		\\ \bottomrule
	\end{tabular}
\end{table}

\subsection{Experimental Setup}
\label{ss.experimental_setup}

The experiments\footnote{The source code is available at \url{https://github.com/gugarosa/opf_speedup}.} were conducted using a $2$-fold cross-validation procedure with $25$ runnings, i.e., each dataset is split into training and testing sets $25$ times. After the split, OPF and Numba-based OPF classifiers are trained using the training set and evaluated over the testing set. Finally, the experimental tables are composed of the training times' mean values\footnote{Note that the prediction time is inferior to the training time and follows the same speed up trend when using Numba.}.
\section{Experimental Results}
\label{s.experiments}

Tables~\ref{t.experiment_a},~\ref{t.experiment_b},~\ref{t.experiment_c},~\ref{t.experiment_d} and~\ref{t.experiment_e} describe the mean OPF training time with and without Numba in seconds over the training sets evaluated by the proposed classifiers. Additionally, the bolded cells are statistically equivalent according to the Wilcoxon signed-rank test. When comparing the OPF + Numba classifier, it is possible to observe that it could achieve, at least once, the best training time across all $22$ datasets. On the other hand, it is important to remark that the Numba-less version almost achieved a similar performance in the smallest datasets, such as Arcene, Lung, ORL, Sonar, Vehicle, and Wine.

\begin{sidewaystable}[!ht]
	\renewcommand{\arraystretch}{2.5}
	\setlength{\tabcolsep}{10pt}
    \centering
	\caption{Mean OPF training time with Numba [without Numba] in seconds and evaluated by $D_1-D_{10}$ classifiers.}
	\vspace*{0.3cm}
    \label{t.experiment_a}
    \scalebox{0.6}{
	\begin{tabular}{lcccccccccc}
		\toprule
		& $\mathbf{D_1}$ & $\mathbf{D_2}$ & $\mathbf{D_3}$ & $\mathbf{D_4}$ & $\mathbf{D_5}$ & $\mathbf{D_6}$ & $\mathbf{D_7}$ & $\mathbf{D_8}$ & $\mathbf{D_9}$ & $\mathbf{D_{10}}$
		\\ \midrule
Arcene & $0.279 \; [\mathbf{0.248}]$ & $0.147 \; [\mathbf{0.126}]$ & $\mathbf{0.105} \; [0.112]$ & $\mathbf{0.264} \; [0.308]$ & $\mathbf{0.180} \; [0.468]$ & $\mathbf{0.146} \; [0.169]$ & $\mathbf{0.191} \; [0.227]$ & $0.348 \; [\mathbf{0.295}]$ & $\mathbf{0.189} \; [0.316]$ & $0.360 \; [\mathbf{0.290}]$
\\
BASEHOCK & $\mathbf{10.837} \; [17.066]$ & $\mathbf{6.717} \; [9.354]$ & $\mathbf{9.820} \; [12.837]$ & $\mathbf{14.453} \; [20.249]$ & $\mathbf{9.546} \; [25.590]$ & $\mathbf{6.458} \; [11.213]$ & $\mathbf{10.440} \; [14.903]$ & $\mathbf{18.699} \; [20.913]$ & $\mathbf{10.285} \; [17.470]$ & $\mathbf{19.158} \; [20.701]$
\\
Caltech101 & $\mathbf{75.663} \; [130.257]$ & $\mathbf{42.436} \; [88.782]$ & $\mathbf{35.922} \; [54.214]$ & $\mathbf{83.210} \; [133.859]$ & $\mathbf{72.659} \; [148.710]$ & $\mathbf{25.605} \; [52.495]$ & $\mathbf{74.080} \; [120.244]$ & $\mathbf{94.568} \; [177.026]$ & $\mathbf{74.693} \; [139.293]$ & $\mathbf{95.136} \; [176.054]$
\\
COIL20 & $\mathbf{2.831} \; [4.806]$ & $\mathbf{1.653} \; [3.351]$ & $\mathbf{1.111} \; [2.028]$ & $\mathbf{3.150} \; [5.690]$ & $\mathbf{2.757} \; [6.506]$ & $\mathbf{1.515} \; [3.232]$ & $\mathbf{2.799} \; [4.819]$ & $\mathbf{3.620} \; [6.943]$ & $\mathbf{2.743} \; [5.565]$ & $\mathbf{3.591} \; [6.833]$
\\
Isolet & $\mathbf{2.010} \; [3.450]$ & $\mathbf{1.560} \; [3.274]$ & $\mathbf{1.320} \; [2.137]$ & $\mathbf{2.879} \; [5.361]$ & $\mathbf{2.544} \; [5.665]$ & $\mathbf{1.305} \; [2.948]$ & $\mathbf{2.675} \; [4.907]$ & $\mathbf{3.292} \; [7.380]$ & $\mathbf{2.516} \; [5.199]$ & $\mathbf{3.307} \; [7.123]$
\\
Lung & $0.163 \; [\mathbf{0.158}]$ & $\mathbf{0.069} \; [0.084]$ & $0.064 \; [\mathbf{0.060}]$ & $\mathbf{0.126} \; [0.164]$ & $\mathbf{0.099} \; [0.228]$ & $\mathbf{0.067} \; [0.101]$ & $\mathbf{0.100} \; [0.135]$ & $\mathbf{0.156} \; [0.181]$ & $\mathbf{0.099} \; [0.172]$ & $\mathbf{0.161} \; [0.174]$
\\
Madelon & $\mathbf{6.670} \; [12.194]$ & $\mathbf{3.923} \; [8.680]$ & $\mathbf{3.588} \; [5.843]$ & $\mathbf{7.208} \; [14.160]$ & $\mathbf{6.586} \; [14.834]$ & $\mathbf{3.610} \; [8.141]$ & $\mathbf{6.566} \; [10.771]$ & $\mathbf{7.978} \; [17.921]$ & $\mathbf{6.616} \; [13.770]$ & $\mathbf{8.010} \; [17.855]$
\\
MPEG7 & $\mathbf{2.259} \; [3.767]$ & $\mathbf{1.311} \; [2.441]$ & $\mathbf{1.366} \; [1.836]$ & $\mathbf{2.493} \; [4.336]$ & $\mathbf{2.125} \; [4.922]$ & $\mathbf{0.734} \; [1.547]$ & $\mathbf{2.174} \; [3.717]$ & $\mathbf{2.812} \; [5.457]$ & $\mathbf{2.221} \; [4.282]$ & $\mathbf{2.854} \; [5.233]$
\\
MPEG7-BAS & $\mathbf{1.620} \; [3.140]$ & $\mathbf{0.891} \; [2.098]$ & $\mathbf{0.899} \; [1.496]$ & $\mathbf{1.623} \; [3.293]$ & $\mathbf{1.544} \; [3.234]$ & $\mathbf{0.836} \; [1.934]$ & $\mathbf{1.540} \; [2.903]$ & $\mathbf{1.747} \; [4.368]$ & $\mathbf{1.553} \; [3.207]$ & $\mathbf{1.819} \; [4.297]$
\\
MPEG7-Fourier & $\mathbf{1.140} \; [2.074]$ & $\mathbf{0.700} \; [1.595]$ & $\mathbf{0.851} \; [1.428]$ & $\mathbf{1.258} \; [2.455]$ & $\mathbf{1.017} \; [1.957]$ & $\mathbf{0.694} \; [1.505]$ & $\mathbf{1.141} \; [2.048]$ & $\mathbf{1.292} \; [3.283]$ & $\mathbf{1.008} \; [1.976]$ & $\mathbf{1.321} \; [3.242]$
\\
Mushrooms & $\mathbf{59.325} \; [109.131]$ & $\mathbf{30.911} \; [117.083]$ & $\mathbf{22.503} \; [45.581]$ & $\mathbf{59.520} \; [106.174]$ & $\mathbf{58.267} \; [117.324]$ & $\mathbf{25.952} \; [62.104]$ & $\mathbf{56.639} \; [103.521]$ & $\mathbf{60.384} \; [158.796]$ & $\mathbf{58.406} \; [119.774]$ & $\mathbf{61.617} \; [139.497]$
\\
NTL-Commercial & $\mathbf{19.183} \; [44.408]$ & $\mathbf{11.853} \; [29.617]$ & $\mathbf{21.157} \; [34.615]$ & $\mathbf{20.328} \; [44.937]$ & $\mathbf{19.862} \; [41.645]$ & $\mathbf{11.289} \; [28.762]$ & $\mathbf{20.975} \; [40.238]$ & $\mathbf{19.932} \; [62.163]$ & $\mathbf{19.730} \; [43.312]$ & $\mathbf{20.838} \; [59.970]$
\\
NTL-Industrial & $\mathbf{8.272} \; [16.656]$ & $\mathbf{5.573} \; [12.593]$ & $\mathbf{8.882} \; [14.277]$ & $\mathbf{8.634} \; [18.788]$ & $\mathbf{8.821} \; [17.174]$ & $\mathbf{5.428} \; [12.204]$ & $\mathbf{9.093} \; [16.777]$ & $\mathbf{8.416} \; [25.055]$ & $\mathbf{8.736} \; [17.695]$ & $\mathbf{9.344} \; [24.360]$
\\
ORL & $\mathbf{0.245} \; [0.333]$ & $\mathbf{0.120} \; [0.200]$ & $\mathbf{0.100} \; [0.156]$ & $\mathbf{0.211} \; [0.352]$ & $\mathbf{0.187} \; [0.396]$ & $\mathbf{0.096} \; [0.173]$ & $\mathbf{0.186} \; [0.300]$ & $\mathbf{0.240} \; [0.428]$ & $\mathbf{0.186} \; [0.349]$ & $\mathbf{0.241} \; [0.424]$
\\
PCMAC & $\mathbf{7.887} \; [12.703]$ & $\mathbf{4.782} \; [7.259]$ & $\mathbf{7.304} \; [10.067]$ & $\mathbf{10.043} \; [15.105]$ & $\mathbf{7.193} \; [18.995]$ & $\mathbf{4.590} \; [8.491]$ & $\mathbf{7.639} \; [11.919]$ & $\mathbf{12.631} \; [15.959]$ & $\mathbf{7.496} \; [13.200]$ & $\mathbf{12.681} \; [15.861]$
\\
Phishing & $\mathbf{111.629} \; [205.797]$ & $\mathbf{55.919} \; [137.076]$ & $\mathbf{42.184} \; [85.402]$ & $\mathbf{102.986} \; [198.750]$ & $\mathbf{108.049} \; [207.306]$ & $\mathbf{50.057} \; [111.415]$ & $\mathbf{102.680} \; [175.067]$ & $\mathbf{107.248} \; [288.288]$ & $\mathbf{107.859} \; [205.874]$ & $\mathbf{108.303} \; [287.348]$
\\
Segment & $\mathbf{5.174} \; [9.694]$ & $\mathbf{2.641} \; [6.299]$ & $\mathbf{2.270} \; [3.669]$ & $\mathbf{4.513} \; [9.774]$ & $\mathbf{4.739} \; [9.090]$ & $\mathbf{2.622} \; [6.256]$ & $\mathbf{4.551} \; [8.969]$ & $\mathbf{4.447} \; [13.710]$ & $\mathbf{4.268} \; [9.571]$ & $\mathbf{4.875} \; [13.561]$
\\
Semeion & $\mathbf{2.592} \; [4.830]$ & $\mathbf{1.426} \; [3.142]$ & $\mathbf{1.016} \; [1.977]$ & $\mathbf{2.679} \; [5.302]$ & $\mathbf{2.425} \; [5.171]$ & $\mathbf{0.802} \; [1.926]$ & $\mathbf{2.476} \; [4.704]$ & $\mathbf{2.802} \; [7.067]$ & $\mathbf{2.465} \; [5.170]$ & $\mathbf{2.874} \; [6.965]$
\\
Sonar & $0.107 \; [\mathbf{0.068}]$ & $\mathbf{0.037} \; [0.054]$ & $\mathbf{0.034} \; [0.036]$ & $\mathbf{0.050} \; [0.079]$ & $\mathbf{0.050} \; [0.076]$ & $\mathbf{0.036} \; [0.053]$ & $\mathbf{0.051} \; [0.074]$ & $\mathbf{0.056} \; [0.105]$ & $\mathbf{0.053} \; [0.077]$ & $\mathbf{0.061} \; [0.102]$
\\
Spambase & $\mathbf{19.150} \; [35.577]$ & $\mathbf{10.497} \; [28.420]$ & $\mathbf{9.941} \; [15.841]$ & $\mathbf{17.648} \; [35.339]$ & $\mathbf{18.898} \; [36.742]$ & $\mathbf{7.356} \; [19.170]$ & $\mathbf{18.507} \; [33.671]$ & $\mathbf{19.127} \; [53.555]$ & $\mathbf{18.822} \; [38.461]$ & $\mathbf{19.728} \; [53.773]$
\\
Vehicle & $\mathbf{0.607} \; [0.974]$ & $\mathbf{0.332} \; [0.746]$ & $\mathbf{0.311} \; [0.518]$ & $\mathbf{0.539} \; [1.201]$ & $\mathbf{0.588} \; [1.123]$ & $\mathbf{0.315} \; [0.721]$ & $\mathbf{0.572} \; [1.058]$ & $\mathbf{0.534} \; [1.586]$ & $\mathbf{0.537} \; [1.105]$ & $\mathbf{0.583} \; [1.525]$
\\
Wine & $0.097 \; [\mathbf{0.057}]$ & $\mathbf{0.029} \; [0.044]$ & $0.027 \; [\mathbf{0.026}]$ & $\mathbf{0.042} \; [0.064]$ & $\mathbf{0.042} \; [0.060]$ & $\mathbf{0.031} \; [0.043]$ & $\mathbf{0.043} \; [0.058]$ & $\mathbf{0.041} \; [0.084]$ & $\mathbf{0.041} \; [0.061]$ & $\mathbf{0.046} \; [0.081]$
\\ \bottomrule
	\end{tabular}}
\end{sidewaystable}

\begin{sidewaystable}[!ht]
	\renewcommand{\arraystretch}{2.5}
	\setlength{\tabcolsep}{10pt}
    \centering
	\caption{Mean OPF training time with Numba [without Numba] in seconds and evaluated by $D_{11}-D_{20}$ classifiers.}
	\vspace*{0.3cm}
    \label{t.experiment_b}
    \scalebox{0.6}{
	\begin{tabular}{lcccccccccc}
		\toprule
		& $\mathbf{D_{11}}$ & $\mathbf{D_{12}}$ & $\mathbf{D_{13}}$ & $\mathbf{D_{14}}$ & $\mathbf{D_{15}}$ & $\mathbf{D_{16}}$ & $\mathbf{D_{17}}$ & $\mathbf{D_{18}}$ & $\mathbf{D_{19}}$ & $\mathbf{D_{20}}$
		\\ \midrule
Arcene & $0.349 \; [\mathbf{0.285}]$ & $\mathbf{0.188} \; [0.241]$ & $0.133 \; [\mathbf{0.120}]$ & $0.121 \; [\mathbf{0.110}]$ & $\mathbf{0.134} \; [0.197]$ & $0.092 \; [\mathbf{0.053}]$ & $\mathbf{0.344} \; [0.992]$ & $\mathbf{0.214} \; [0.255]$ & $\mathbf{0.322} \; [0.373]$ & $1.437 \; [\mathbf{0.516}]$
\\
BASEHOCK & $\mathbf{18.363} \; [20.677]$ & $\mathbf{10.192} \; [16.157]$ & $\mathbf{6.745} \; [8.597]$ & $\mathbf{6.153} \; [7.933]$ & $\mathbf{6.525} \; [11.983]$ & $\mathbf{2.723} \; [3.356]$ & $\mathbf{16.123} \; [63.284]$ & $\mathbf{9.986} \; [15.971]$ & $\mathbf{25.154} \; [26.400]$ & $57.589 \; [\mathbf{37.562}]$
\\
Caltech101 & $\mathbf{94.792} \; [172.639]$ & $\mathbf{75.634} \; [133.664]$ & $\mathbf{42.829} \; [81.069]$ & $\mathbf{25.164} \; [48.095]$ & $\mathbf{43.369} \; [87.293]$ & $\mathbf{21.001} \; [29.962]$ & $\mathbf{84.236} \; [520.868]$ & $\mathbf{45.852} \; [112.818]$ & $240.621 \; [\mathbf{213.830}]$ & $\mathbf{176.646} \; [181.138]$
\\
COIL20 & $\mathbf{3.553} \; [6.797]$ & $\mathbf{2.776} \; [5.249]$ & $\mathbf{1.626} \; [3.138]$ & $\mathbf{0.817} \; [1.532]$ & $\mathbf{1.640} \; [3.472]$ & $\mathbf{0.882} \; [1.104]$ & $\mathbf{3.344} \; [19.624]$ & $\mathbf{2.123} \; [4.637]$ & $9.455 \; [\mathbf{8.385}]$ & $10.066 \; [\mathbf{6.008}]$
\\
Isolet & $\mathbf{3.288} \; [7.003]$ & $\mathbf{2.562} \; [4.962]$ & $\mathbf{1.550} \; [3.171]$ & $\mathbf{0.791} \; [1.658]$ & $\mathbf{1.509} \; [3.267]$ & $\mathbf{0.899} \; [1.059]$ & $\mathbf{2.901} \; [18.191]$ & $\mathbf{1.857} \; [4.405]$ & $10.035 \; [\mathbf{8.526}]$ & $7.493 \; [\mathbf{5.497}]$
\\
Lung & $\mathbf{0.158} \; [0.177]$ & $\mathbf{0.100} \; [0.146]$ & $\mathbf{0.068} \; [0.081]$ & $\mathbf{0.039} \; [0.042]$ & $\mathbf{0.067} \; [0.106]$ & $0.037 \; [\mathbf{0.030}]$ & $\mathbf{0.154} \; [0.543]$ & $\mathbf{0.099} \; [0.137]$ & $0.232 \; [\mathbf{0.219}]$ & $0.618 \; [\mathbf{0.316}]$
\\
Madelon & $\mathbf{7.974} \; [17.494]$ & $\mathbf{6.631} \; [13.187]$ & $\mathbf{3.954} \; [7.783]$ & $\mathbf{3.978} \; [7.826]$ & $\mathbf{3.849} \; [8.079]$ & $\mathbf{2.464} \; [2.896]$ & $\mathbf{7.100} \; [46.282]$ & $\mathbf{4.562} \; [10.801]$ & $25.576 \; [\mathbf{21.693}]$ & $\mathbf{17.016} \; [20.584]$
\\
MPEG7 & $\mathbf{2.874} \; [5.231]$ & $\mathbf{2.217} \; [4.101]$ & $\mathbf{1.305} \; [2.366]$ & $\mathbf{0.749} \; [1.401]$ & $\mathbf{1.298} \; [2.667]$ & $\mathbf{0.677} \; [0.808]$ & $\mathbf{2.575} \; [14.874]$ & $\mathbf{1.658} \; [3.461]$ & $7.741 \; [\mathbf{6.430}]$ & $6.720 \; [\mathbf{5.672}]$
\\
MPEG7-BAS & $\mathbf{1.749} \; [4.238]$ & $\mathbf{1.566} \; [3.202]$ & $\mathbf{0.885} \; [1.913]$ & $\mathbf{0.532} \; [1.213]$ & $\mathbf{0.892} \; [1.911]$ & $\mathbf{0.588} \; [0.717]$ & $\mathbf{1.502} \; [11.025]$ & $\mathbf{0.946} \; [2.538]$ & $6.164 \; [\mathbf{4.917}]$ & $\mathbf{2.543} \; [3.325]$
\\
MPEG7-Fourier & $\mathbf{1.325} \; [3.230]$ & $\mathbf{1.007} \; [2.006]$ & $\mathbf{0.684} \; [1.474]$ & $\mathbf{0.518} \; [1.158]$ & $\mathbf{0.667} \; [1.403]$ & $\mathbf{0.446} \; [0.550]$ & $\mathbf{1.064} \; [8.266]$ & $\mathbf{0.659} \; [1.841]$ & $4.667 \; [\mathbf{4.031}]$ & $\mathbf{1.733} \; [2.106]$
\\
Mushrooms & $\mathbf{62.466} \; [155.330]$ & $\mathbf{59.226} \; [119.482]$ & $\mathbf{31.423} \; [79.656]$ & $\mathbf{23.768} \; [53.317]$ & $\mathbf{31.828} \; [71.157]$ & $\mathbf{22.691} \; [30.154]$ & $\mathbf{54.939} \; [407.768]$ & $\mathbf{31.064} \; [85.818]$ & $215.523 \; [\mathbf{185.357}]$ & $\mathbf{72.644} \; [112.182]$
\\
NTL-Commercial & $\mathbf{20.473} \; [59.755]$ & $\mathbf{19.424} \; [44.084]$ & $\mathbf{11.426} \; [27.521]$ & $\mathbf{12.171} \; [27.996]$ & $\mathbf{11.058} \; [26.404]$ & $\mathbf{8.307} \; [10.744]$ & $\mathbf{20.128} \; [153.952]$ & $\mathbf{10.949} \; [34.998]$ & $82.519 \; [\mathbf{72.154}]$ & $\mathbf{20.328} \; [39.648]$
\\
NTL-Industrial & $\mathbf{9.676} \; [24.751]$ & $\mathbf{8.880} \; [17.976]$ & $\mathbf{5.214} \; [11.691]$ & $\mathbf{5.070} \; [11.535]$ & $\mathbf{4.846} \; [11.185]$ & $\mathbf{3.748} \; [4.562]$ & $\mathbf{8.595} \; [63.656]$ & $\mathbf{5.109} \; [14.518]$ & $37.240 \; [\mathbf{29.655}]$ & $\mathbf{9.477} \; [16.238]$
\\
ORL & $\mathbf{0.241} \; [0.416]$ & $\mathbf{0.186} \; [0.315]$ & $\mathbf{0.118} \; [0.191]$ & $\mathbf{0.074} \; [0.119]$ & $\mathbf{0.116} \; [0.218]$ & $\mathbf{0.082} \; [0.086]$ & $\mathbf{0.216} \; [1.202]$ & $\mathbf{0.147} \; [0.285]$ & $0.580 \; [\mathbf{0.513}]$ & $0.681 \; [\mathbf{0.427}]$
\\
PCMAC & $\mathbf{12.534} \; [15.814]$ & $\mathbf{7.600} \; [12.687]$ & $\mathbf{4.806} \; [6.897]$ & $\mathbf{4.563} \; [6.502]$ & $\mathbf{4.712} \; [9.010]$ & $\mathbf{2.198} \; [2.632]$ & $\mathbf{11.175} \; [48.489]$ & $\mathbf{6.937} \; [11.824]$ & $\mathbf{20.360} \; [20.541]$ & $36.933 \; [\mathbf{27.081}]$
\\
Phishing & $\mathbf{107.815} \; [286.369]$ & $\mathbf{109.512} \; [220.978]$ & $\mathbf{58.793} \; [173.069]$ & $\mathbf{46.914} \; [156.068]$ & $\mathbf{58.870} \; [122.099]$ & $\mathbf{42.126} \; [51.195]$ & $\mathbf{100.449} \; [769.463]$ & $\mathbf{55.982} \; [178.092]$ & $404.487 \; [\mathbf{357.772}]$ & $\mathbf{121.653} \; [211.029]$
\\
Segment & $\mathbf{4.692} \; [13.238]$ & $\mathbf{4.271} \; [9.742]$ & $\mathbf{2.621} \; [6.164]$ & $\mathbf{1.103} \; [2.958]$ & $\mathbf{2.454} \; [5.989]$ & $\mathbf{1.636} \; [1.960]$ & $\mathbf{4.432} \; [34.023]$ & $\mathbf{2.647} \; [7.791]$ & $19.829 \; [\mathbf{15.890}]$ & $\mathbf{5.177} \; [8.683]$
\\
Semeion & $\mathbf{2.872} \; [6.772]$ & $\mathbf{2.555} \; [5.151]$ & $\mathbf{1.422} \; [3.114]$ & $\mathbf{0.724} \; [1.645]$ & $\mathbf{1.438} \; [3.163]$ & $\mathbf{1.009} \; [1.214]$ & $\mathbf{2.554} \; [18.097]$ & $\mathbf{1.584} \; [4.221]$ & $10.215 \; [\mathbf{8.159}]$ & $\mathbf{4.558} \; [5.158]$
\\
Sonar & $\mathbf{0.064} \; [0.099]$ & $\mathbf{0.062} \; [0.077]$ & $\mathbf{0.038} \; [0.050]$ & $\mathbf{0.036} \; [0.047]$ & $\mathbf{0.034} \; [0.049]$ & $0.031 \; [\mathbf{0.023}]$ & $\mathbf{0.052} \; [0.261]$ & $\mathbf{0.036} \; [0.064]$ & $0.151 \; [\mathbf{0.119}]$ & $\mathbf{0.061} \; [0.075]$
\\
Spambase & $\mathbf{19.625} \; [51.123]$ & $\mathbf{19.001} \; [38.895]$ & $\mathbf{10.966} \; [30.180]$ & $\mathbf{9.854} \; [27.322]$ & $\mathbf{9.521} \; [23.222]$ & $\mathbf{7.935} \; [10.090]$ & $\mathbf{17.179} \; [131.594]$ & $\mathbf{9.703} \; [31.520]$ & $72.230 \; [\mathbf{63.171}]$ & $\mathbf{20.442} \; [35.314]$
\\
Vehicle & $\mathbf{0.571} \; [1.494]$ & $\mathbf{0.525} \; [1.094]$ & $\mathbf{0.330} \; [0.704]$ & $\mathbf{0.270} \; [0.642]$ & $\mathbf{0.303} \; [0.692]$ & $\mathbf{0.259} \; [0.300]$ & $\mathbf{0.533} \; [4.000]$ & $\mathbf{0.333} \; [0.897]$ & $2.312 \; [\mathbf{1.836}]$ & $\mathbf{0.592} \; [1.020]$
\\
Wine & $\mathbf{0.043} \; [0.080]$ & $\mathbf{0.042} \; [0.061]$ & $\mathbf{0.031} \; [0.040]$ & $\mathbf{0.019} \; [0.022]$ & $\mathbf{0.030} \; [0.039]$ & $0.027 \; [\mathbf{0.017}]$ & $\mathbf{0.043} \; [0.199]$ & $\mathbf{0.030} \; [0.050]$ & $0.120 \; [\mathbf{0.093}]$ & $\mathbf{0.044} \; [0.055]$
\\ \bottomrule
	\end{tabular}}
\end{sidewaystable}

\begin{sidewaystable}[!ht]
	\renewcommand{\arraystretch}{2.5}
	\setlength{\tabcolsep}{10pt}
    \centering
	\caption{Mean OPF training time with Numba [without Numba] in seconds and evaluated by $D_{21}-D_{30}$ classifiers.}
	\vspace*{0.3cm}
    \label{t.experiment_c}
    \scalebox{0.6}{
	\begin{tabular}{lcccccccccc}
		\toprule
		& $\mathbf{D_{21}}$ & $\mathbf{D_{22}}$ & $\mathbf{D_{23}}$ & $\mathbf{D_{24}}$ & $\mathbf{D_{25}}$ & $\mathbf{D_{26}}$ & $\mathbf{D_{27}}$ & $\mathbf{D_{28}}$ & $\mathbf{D_{29}}$ & $\mathbf{D_{30}}$
		\\ \midrule
Arcene & $2.968 \; [\mathbf{0.810}]$ & $3.233 \; [\mathbf{1.180}]$ & $1.469 \; [\mathbf{0.593}]$ & $\mathbf{0.270} \; [0.306]$ & $1.436 \; [\mathbf{0.461}]$ & $0.136 \; [\mathbf{0.129}]$ & $0.136 \; [\mathbf{0.126}]$ & $1.198 \; [\mathbf{1.084}]$ & $\mathbf{0.134} \; [0.193]$ & $\mathbf{0.194} \; [0.233]$
\\
BASEHOCK & $148.520 \; [\mathbf{47.839}]$ & $110.239 \; [\mathbf{96.700}]$ & $53.827 \; [\mathbf{48.583}]$ & $\mathbf{14.853} \; [20.640]$ & $55.234 \; [\mathbf{33.879}]$ & $\mathbf{6.718} \; [9.013]$ & $\mathbf{6.732} \; [8.684]$ & $\mathbf{21.835} \; [31.538]$ & $\mathbf{6.520} \; [11.727]$ & $\mathbf{9.729} \; [14.489]$
\\
Caltech101 & $\mathbf{306.061} \; [332.199]$ & $\mathbf{323.904} \; [340.791]$ & $\mathbf{122.739} \; [148.685]$ & $\mathbf{85.175} \; [158.874]$ & $\mathbf{122.843} \; [126.355]$ & $\mathbf{43.393} \; [85.736]$ & $\mathbf{43.041} \; [82.862]$ & $\mathbf{119.278} \; [169.153]$ & $\mathbf{42.718} \; [84.270]$ & $\mathbf{46.333} \; [103.939]$
\\
COIL20 & $23.035 \; [\mathbf{10.325}]$ & $18.644 \; [\mathbf{11.949}]$ & $7.050 \; [\mathbf{4.784}]$ & $\mathbf{3.209} \; [5.780]$ & $9.206 \; [\mathbf{5.024}]$ & $\mathbf{1.655} \; [3.308]$ & $\mathbf{1.675} \; [3.224]$ & $7.219 \; [\mathbf{6.451}]$ & $\mathbf{1.640} \; [3.232]$ & $\mathbf{1.911} \; [4.049]$
\\
Isolet & $16.448 \; [\mathbf{10.705}]$ & $13.645 \; [\mathbf{11.668}]$ & $4.693 \; [\mathbf{4.286}]$ & $\mathbf{2.940} \; [5.947]$ & $5.994 \; [\mathbf{4.143}]$ & $\mathbf{1.587} \; [3.433]$ & $\mathbf{1.591} \; [3.403]$ & $5.720 \; [\mathbf{5.715}]$ & $\mathbf{1.533} \; [3.167]$ & $\mathbf{1.684} \; [4.089]$
\\
Lung & $1.429 \; [\mathbf{0.369}]$ & $1.130 \; [\mathbf{0.789}]$ & $0.470 \; [\mathbf{0.337}]$ & $\mathbf{0.128} \; [0.168]$ & $0.597 \; [\mathbf{0.288}]$ & $\mathbf{0.068} \; [0.084]$ & $\mathbf{0.068} \; [0.083]$ & $0.475 \; [\mathbf{0.426}]$ & $\mathbf{0.068} \; [0.103]$ & $\mathbf{0.087} \; [0.126]$
\\
Madelon & $35.106 \; [\mathbf{22.944}]$ & $\mathbf{24.608} \; [38.730]$ & $\mathbf{9.589} \; [13.331]$ & $\mathbf{7.264} \; [14.420]$ & $\mathbf{10.426} \; [12.216]$ & $\mathbf{3.924} \; [8.550]$ & $\mathbf{3.911} \; [8.052]$ & $\mathbf{12.503} \; [18.313]$ & $\mathbf{3.842} \; [7.585]$ & $\mathbf{4.136} \; [9.641]$
\\
MPEG7 & $14.292 \; [\mathbf{9.143}]$ & $12.191 \; [\mathbf{11.406}]$ & $5.521 \; [\mathbf{5.288}]$ & $\mathbf{2.548} \; [4.430]$ & $5.463 \; [\mathbf{4.331}]$ & $\mathbf{1.313} \; [2.485]$ & $\mathbf{1.319} \; [2.401]$ & $\mathbf{4.043} \; [4.661]$ & $\mathbf{1.302} \; [2.456]$ & $\mathbf{1.465} \; [3.115]$
\\
MPEG7-BAS & $\mathbf{4.151} \; [4.766]$ & $\mathbf{3.607} \; [5.920]$ & $\mathbf{1.577} \; [2.349]$ & $\mathbf{1.636} \; [3.259]$ & $\mathbf{1.637} \; [1.999]$ & $\mathbf{0.904} \; [2.025]$ & $\mathbf{0.907} \; [1.941]$ & $\mathbf{1.730} \; [2.994]$ & $\mathbf{0.896} \; [1.745]$ & $\mathbf{0.845} \; [2.232]$
\\
MPEG7-Fourier & $\mathbf{2.627} \; [3.306]$ & $\mathbf{2.519} \; [3.466]$ & $\mathbf{1.441} \; [1.891]$ & $\mathbf{1.265} \; [2.516]$ & $\mathbf{1.488} \; [1.629]$ & $\mathbf{0.708} \; [1.636]$ & $\mathbf{0.710} \; [1.594]$ & $\mathbf{1.057} \; [2.150]$ & $\mathbf{0.658} \; [1.365]$ & $\mathbf{0.583} \; [1.600]$
\\
Mushrooms & $\mathbf{104.102} \; [177.517]$ & $\mathbf{98.558} \; [186.505]$ & $\mathbf{72.402} \; [124.696]$ & $\mathbf{60.902} \; [124.279]$ & $\mathbf{72.704} \; [108.129]$ & $\mathbf{31.571} \; [87.446]$ & $\mathbf{31.783} \; [74.253]$ & $\mathbf{42.871} \; [96.194]$ & $\mathbf{31.953} \; [64.520]$ & $\mathbf{30.398} \; [82.343]$
\\
NTL-Commercial & $\mathbf{21.399} \; [61.641]$ & $\mathbf{21.806} \; [62.784]$ & $\mathbf{11.864} \; [26.920]$ & $\mathbf{20.178} \; [46.130]$ & $\mathbf{11.908} \; [23.414]$ & $\mathbf{11.790} \; [28.075]$ & $\mathbf{11.915} \; [29.343]$ & $\mathbf{12.000} \; [31.951]$ & $\mathbf{11.409} \; [24.809]$ & $\mathbf{11.103} \; [30.982]$
\\
NTL-Industrial & $\mathbf{9.895} \; [25.265]$ & $\mathbf{10.197} \; [25.675]$ & $\mathbf{5.457} \; [10.166]$ & $\mathbf{9.173} \; [19.126]$ & $\mathbf{5.382} \; [8.745]$ & $\mathbf{5.407} \; [11.436]$ & $\mathbf{5.266} \; [12.052]$ & $\mathbf{5.604} \; [13.522]$ & $\mathbf{5.122} \; [10.649]$ & $\mathbf{4.889} \; [13.120]$
\\
ORL & $1.420 \; [\mathbf{0.632}]$ & $1.141 \; [\mathbf{0.889}]$ & $0.426 \; [\mathbf{0.348}]$ & $\mathbf{0.219} \; [0.357]$ & $0.504 \; [\mathbf{0.275}]$ & $\mathbf{0.119} \; [0.203]$ & $\mathbf{0.119} \; [0.198]$ & $0.523 \; [\mathbf{0.444}]$ & $\mathbf{0.117} \; [0.202]$ & $\mathbf{0.129} \; [0.256]$
\\
PCMAC & $97.353 \; [\mathbf{34.230}]$ & $70.093 \; [\mathbf{67.195}]$ & $34.939 \; [\mathbf{34.237}]$ & $\mathbf{10.308} \; [15.282]$ & $35.727 \; [\mathbf{24.788}]$ & $\mathbf{4.854} \; [7.103]$ & $\mathbf{4.806} \; [7.041]$ & $\mathbf{13.694} \; [21.057]$ & $\mathbf{4.726} \; [8.698]$ & $\mathbf{6.796} \; [10.694]$
\\
Phishing & $\mathbf{152.954} \; [342.190]$ & $\mathbf{153.936} \; [347.482]$ & $\mathbf{121.545} \; [234.939]$ & $\mathbf{108.651} \; [230.882]$ & $\mathbf{125.343} \; [199.707]$ & $\mathbf{59.618} \; [205.591]$ & $\mathbf{60.487} \; [138.205]$ & $\mathbf{74.331} \; [169.863]$ & $\mathbf{59.876} \; [125.598]$ & $\mathbf{57.194} \; [144.318]$
\\
Segment & $\mathbf{5.877} \; [13.565]$ & $\mathbf{5.883} \; [13.832]$ & $\mathbf{2.997} \; [5.370]$ & $\mathbf{4.764} \; [10.174]$ & $\mathbf{3.174} \; [4.928]$ & $\mathbf{2.601} \; [6.418]$ & $\mathbf{2.615} \; [6.220]$ & $\mathbf{3.011} \; [6.990]$ & $\mathbf{2.547} \; [5.431]$ & $\mathbf{2.516} \; [6.972]$
\\
Semeion & $\mathbf{7.465} \; [8.772]$ & $\mathbf{7.085} \; [8.959]$ & $\mathbf{4.206} \; [5.292]$ & $\mathbf{2.751} \; [5.376]$ & $\mathbf{4.295} \; [4.543]$ & $\mathbf{1.474} \; [3.279]$ & $\mathbf{1.483} \; [3.222]$ & $\mathbf{2.781} \; [4.415]$ & $\mathbf{1.430} \; [2.879]$ & $\mathbf{1.437} \; [3.725]$
\\
Sonar & $\mathbf{0.076} \; [0.112]$ & $\mathbf{0.077} \; [0.115]$ & $\mathbf{0.051} \; [0.063]$ & $\mathbf{0.054} \; [0.081]$ & $\mathbf{0.056} \; [0.060]$ & $\mathbf{0.037} \; [0.055]$ & $\mathbf{0.038} \; [0.053]$ & $\mathbf{0.047} \; [0.062]$ & $\mathbf{0.036} \; [0.048]$ & $\mathbf{0.032} \; [0.059]$
\\
Spambase & $\mathbf{24.528} \; [55.754]$ & $\mathbf{24.351} \; [57.123]$ & $\mathbf{10.502} \; [22.067]$ & $\mathbf{17.593} \; [40.792]$ & $\mathbf{11.009} \; [18.301]$ & $\mathbf{11.040} \; [29.151]$ & $\mathbf{10.999} \; [24.250]$ & $\mathbf{11.698} \; [28.888]$ & $\mathbf{9.681} \; [22.034]$ & $\mathbf{9.685} \; [28.027]$
\\
Vehicle & $\mathbf{0.689} \; [1.586]$ & $\mathbf{0.684} \; [1.617]$ & $\mathbf{0.372} \; [0.675]$ & $\mathbf{0.584} \; [1.209]$ & $\mathbf{0.398} \; [0.638]$ & $\mathbf{0.319} \; [0.786]$ & $\mathbf{0.314} \; [0.747]$ & $\mathbf{0.381} \; [0.842]$ & $\mathbf{0.310} \; [0.655]$ & $\mathbf{0.302} \; [0.798]$
\\
Wine & $\mathbf{0.047} \; [0.083]$ & $\mathbf{0.048} \; [0.084]$ & $\mathbf{0.033} \; [0.041]$ & $\mathbf{0.043} \; [0.062]$ & $\mathbf{0.038} \; [0.043]$ & $\mathbf{0.029} \; [0.043]$ & $\mathbf{0.029} \; [0.042]$ & $\mathbf{0.033} \; [0.045]$ & $\mathbf{0.028} \; [0.037]$ & $\mathbf{0.027} \; [0.045]$
\\ \bottomrule
	\end{tabular}}
\end{sidewaystable}

\begin{sidewaystable}[!ht]
	\renewcommand{\arraystretch}{2.5}
	\setlength{\tabcolsep}{10pt}
    \centering
	\caption{Mean OPF training time with Numba [without Numba] in seconds and evaluated by $D_{31}-D_{40}$ classifiers.}
	\vspace*{0.3cm}
    \label{t.experiment_d}
    \scalebox{0.6}{
	\begin{tabular}{lcccccccccc}
		\toprule
		& $\mathbf{D_{31}}$ & $\mathbf{D_{32}}$ & $\mathbf{D_{33}}$ & $\mathbf{D_{34}}$ & $\mathbf{D_{35}}$ & $\mathbf{D_{36}}$ & $\mathbf{D_{37}}$ & $\mathbf{D_{38}}$ & $\mathbf{D_{39}}$ & $\mathbf{D_{40}}$
		\\ \midrule
Arcene & $\mathbf{0.297} \; [0.334]$ & $\mathbf{0.211} \; [0.222]$ & $\mathbf{0.293} \; [0.333]$ & $\mathbf{0.185} \; [0.198]$ & $\mathbf{0.135} \; [0.196]$ & $\mathbf{0.183} \; [0.196]$ & $\mathbf{0.196} \; [0.232]$ & $\mathbf{0.270} \; [0.308]$ & $\mathbf{0.191} \; [0.228]$ & $\mathbf{0.215} \; [0.226]$
\\
BASEHOCK & $\mathbf{16.151} \; [23.490]$ & $\mathbf{10.912} \; [16.149]$ & $\mathbf{14.978} \; [21.856]$ & $\mathbf{10.671} \; [14.207]$ & $\mathbf{6.563} \; [11.952]$ & $\mathbf{10.049} \; [13.366]$ & $\mathbf{10.504} \; [15.450]$ & $\mathbf{14.925} \; [20.999]$ & $\mathbf{10.459} \; [15.324]$ & $\mathbf{10.168} \; [14.555]$
\\
Caltech101 & $\mathbf{86.968} \; [192.566]$ & $\mathbf{77.841} \; [144.342]$ & $\mathbf{65.101} \; [147.405]$ & $\mathbf{65.465} \; [101.408]$ & $\mathbf{42.576} \; [85.306]$ & $\mathbf{61.647} \; [95.164]$ & $\mathbf{74.297} \; [127.586]$ & $\mathbf{85.631} \; [164.714]$ & $\mathbf{75.495} \; [119.165]$ & $\mathbf{46.008} \; [101.808]$
\\
COIL20 & $\mathbf{3.257} \; [6.807]$ & $\mathbf{2.822} \; [5.473]$ & $\mathbf{2.295} \; [4.902]$ & $\mathbf{1.787} \; [2.929]$ & $\mathbf{1.648} \; [3.346]$ & $\mathbf{2.575} \; [4.354]$ & $\mathbf{2.710} \; [4.925]$ & $\mathbf{3.168} \; [5.778]$ & $\mathbf{2.730} \; [4.776]$ & $\mathbf{2.108} \; [4.054]$
\\
Isolet & $\mathbf{2.264} \; [5.388]$ & $\mathbf{2.885} \; [6.173]$ & $\mathbf{2.687} \; [6.299]$ & $\mathbf{1.927} \; [2.794]$ & $\mathbf{1.541} \; [3.303]$ & $\mathbf{1.916} \; [3.278]$ & $\mathbf{2.745} \; [5.143]$ & $\mathbf{3.018} \; [5.955]$ & $\mathbf{2.822} \; [5.139]$ & $\mathbf{1.892} \; [4.001]$
\\
Lung & $\mathbf{0.137} \; [0.195]$ & $\mathbf{0.108} \; [0.141]$ & $\mathbf{0.125} \; [0.176]$ & $\mathbf{0.094} \; [0.115]$ & $\mathbf{0.069} \; [0.104]$ & $\mathbf{0.099} \; [0.118]$ & $\mathbf{0.102} \; [0.135]$ & $\mathbf{0.132} \; [0.167]$ & $\mathbf{0.102} \; [0.133]$ & $\mathbf{0.101} \; [0.123]$
\\
Madelon & $\mathbf{7.588} \; [17.772]$ & $\mathbf{6.909} \; [14.104]$ & $\mathbf{7.648} \; [17.830]$ & $\mathbf{6.587} \; [11.316]$ & $\mathbf{3.925} \; [8.094]$ & $\mathbf{6.683} \; [11.669]$ & $\mathbf{6.642} \; [12.746]$ & $\mathbf{7.277} \; [14.688]$ & $\mathbf{6.636} \; [12.289]$ & $\mathbf{4.556} \; [9.823]$
\\
MPEG7 & $\mathbf{2.623} \; [5.300]$ & $\mathbf{2.338} \; [4.274]$ & $\mathbf{1.929} \; [3.878]$ & $\mathbf{1.919} \; [3.069]$ & $\mathbf{1.299} \; [2.553]$ & $\mathbf{1.868} \; [2.948]$ & $\mathbf{2.215} \; [3.774]$ & $\mathbf{2.549} \; [4.429]$ & $\mathbf{2.201} \; [3.676]$ & $\mathbf{1.635} \; [3.000]$
\\
MPEG7-BAS & $\mathbf{1.646} \; [4.220]$ & $\mathbf{1.626} \; [3.384]$ & $\mathbf{1.663} \; [4.200]$ & $\mathbf{1.525} \; [2.670]$ & $\mathbf{0.893} \; [1.823]$ & $\mathbf{1.527} \; [2.715]$ & $\mathbf{1.530} \; [2.939]$ & $\mathbf{1.621} \; [3.320]$ & $\mathbf{1.527} \; [2.886]$ & $\mathbf{0.941} \; [2.232]$
\\
MPEG7-Fourier & $\mathbf{1.151} \; [3.005]$ & $\mathbf{1.320} \; [2.696]$ & $\mathbf{1.089} \; [2.828]$ & $\mathbf{1.097} \; [1.882]$ & $\mathbf{0.658} \; [1.336]$ & $\mathbf{1.023} \; [1.738]$ & $\mathbf{1.129} \; [2.112]$ & $\mathbf{1.257} \; [2.480]$ & $\mathbf{1.158} \; [2.050]$ & $\mathbf{0.659} \; [1.512]$
\\
Mushrooms & $\mathbf{61.784} \; [158.792]$ & $\mathbf{60.730} \; [194.833]$ & $\mathbf{62.668} \; [159.479]$ & $\mathbf{58.994} \; [102.934]$ & $\mathbf{32.257} \; [66.457]$ & $\mathbf{58.541} \; [100.580]$ & $\mathbf{57.492} \; [106.881]$ & $\mathbf{61.011} \; [124.374]$ & $\mathbf{57.754} \; [104.905]$ & $\mathbf{30.549} \; [80.406]$
\\
NTL-Commercial & $\mathbf{20.065} \; [59.346]$ & $\mathbf{21.840} \; [49.952]$ & $\mathbf{20.140} \; [59.667]$ & $\mathbf{19.727} \; [38.062]$ & $\mathbf{11.337} \; [25.609]$ & $\mathbf{19.564} \; [38.299]$ & $\mathbf{20.327} \; [41.427]$ & $\mathbf{20.602} \; [46.019]$ & $\mathbf{20.320} \; [40.665]$ & $\mathbf{11.086} \; [30.591]$
\\
NTL-Industrial & $\mathbf{8.393} \; [24.721]$ & $\mathbf{9.535} \; [20.864]$ & $\mathbf{9.243} \; [24.772]$ & $\mathbf{8.343} \; [16.225]$ & $\mathbf{5.285} \; [10.955]$ & $\mathbf{9.157} \; [16.127]$ & $\mathbf{9.021} \; [17.149]$ & $\mathbf{9.414} \; [19.453]$ & $\mathbf{9.113} \; [16.863]$ & $\mathbf{5.154} \; [12.777]$
\\
ORL & $\mathbf{0.219} \; [0.429]$ & $\mathbf{0.196} \; [0.351]$ & $\mathbf{0.225} \; [0.444]$ & $\mathbf{0.182} \; [0.276]$ & $\mathbf{0.118} \; [0.204]$ & $\mathbf{0.177} \; [0.265]$ & $\mathbf{0.185} \; [0.295]$ & $\mathbf{0.214} \; [0.356]$ & $\mathbf{0.186} \; [0.294]$ & $\mathbf{0.147} \; [0.249]$
\\
PCMAC & $\mathbf{11.025} \; [17.712]$ & $\mathbf{8.098} \; [12.500]$ & $\mathbf{10.701} \; [16.988]$ & $\mathbf{7.619} \; [10.458]$ & $\mathbf{4.741} \; [8.749]$ & $\mathbf{7.494} \; [10.291]$ & $\mathbf{7.658} \; [12.114]$ & $\mathbf{10.379} \; [15.375]$ & $\mathbf{7.687} \; [12.028]$ & $\mathbf{7.255} \; [10.682]$
\\
Phishing & $\mathbf{118.203} \; [303.478]$ & $\mathbf{117.513} \; [242.081]$ & $\mathbf{119.009} \; [300.492]$ & $\mathbf{112.114} \; [177.669]$ & $\mathbf{60.404} \; [127.375]$ & $\mathbf{113.204} \; [195.746]$ & $\mathbf{113.013} \; [206.874]$ & $\mathbf{112.281} \; [232.312]$ & $\mathbf{111.652} \; [201.835]$ & $\mathbf{57.042} \; [154.125]$
\\
Segment & $\mathbf{4.413} \; [13.048]$ & $\mathbf{4.826} \; [10.775]$ & $\mathbf{4.831} \; [13.121]$ & $\mathbf{4.315} \; [8.497]$ & $\mathbf{2.679} \; [5.683]$ & $\mathbf{4.806} \; [8.649]$ & $\mathbf{4.718} \; [9.237]$ & $\mathbf{4.838} \; [10.376]$ & $\mathbf{4.760} \; [9.027]$ & $\mathbf{2.702} \; [6.800]$
\\
Semeion & $\mathbf{2.696} \; [6.712]$ & $\mathbf{2.631} \; [5.295]$ & $\mathbf{2.513} \; [6.190]$ & $\mathbf{2.302} \; [3.617]$ & $\mathbf{1.434} \; [2.937]$ & $\mathbf{2.313} \; [4.029]$ & $\mathbf{2.457} \; [4.757]$ & $\mathbf{2.670} \; [5.415]$ & $\mathbf{2.484} \; [4.609]$ & $\mathbf{1.560} \; [3.599]$
\\
Sonar & $\mathbf{0.053} \; [0.101]$ & $\mathbf{0.058} \; [0.088]$ & $\mathbf{0.054} \; [0.103]$ & $\mathbf{0.050} \; [0.068]$ & $\mathbf{0.036} \; [0.049]$ & $\mathbf{0.050} \; [0.069]$ & $\mathbf{0.052} \; [0.075]$ & $\mathbf{0.055} \; [0.083]$ & $\mathbf{0.055} \; [0.075]$ & $\mathbf{0.039} \; [0.058]$
\\
Spambase & $\mathbf{19.068} \; [53.409]$ & $\mathbf{20.111} \; [66.299]$ & $\mathbf{19.077} \; [52.703]$ & $\mathbf{18.010} \; [34.040]$ & $\mathbf{9.814} \; [22.978]$ & $\mathbf{17.829} \; [33.881]$ & $\mathbf{19.029} \; [36.834]$ & $\mathbf{18.131} \; [41.479]$ & $\mathbf{18.949} \; [35.677]$ & $\mathbf{9.790} \; [26.838]$
\\
Vehicle & $\mathbf{0.500} \; [1.456]$ & $\mathbf{0.558} \; [1.244]$ & $\mathbf{0.585} \; [1.512]$ & $\mathbf{0.558} \; [0.970]$ & $\mathbf{0.339} \; [0.649]$ & $\mathbf{0.555} \; [0.951]$ & $\mathbf{0.569} \; [1.052]$ & $\mathbf{0.594} \; [1.178]$ & $\mathbf{0.564} \; [1.030]$ & $\mathbf{0.335} \; [0.785]$
\\
Wine & $\mathbf{0.041} \; [0.080]$ & $\mathbf{0.046} \; [0.066]$ & $\mathbf{0.044} \; [0.079]$ & $\mathbf{0.042} \; [0.054]$ & $\mathbf{0.030} \; [0.038]$ & $\mathbf{0.043} \; [0.054]$ & $\mathbf{0.042} \; [0.058]$ & $\mathbf{0.043} \; [0.062]$ & $\mathbf{0.042} \; [0.056]$ & $\mathbf{0.030} \; [0.044]$
\\ \bottomrule
	\end{tabular}}
\end{sidewaystable}

\begin{sidewaystable}[!ht]
	\renewcommand{\arraystretch}{2.5}
	\setlength{\tabcolsep}{10pt}
    \centering
	\caption{Mean OPF training time with Numba [without Numba] in seconds and evaluated by $D_{41}-D_{47}$ classifiers.}
	\vspace*{0.3cm}
    \label{t.experiment_e}
    \scalebox{0.65}{
	\begin{tabular}{lccccccc}
		\toprule
		& $\mathbf{D_{41}}$ & $\mathbf{D_{42}}$ & $\mathbf{D_{43}}$ & $\mathbf{D_{44}}$ & $\mathbf{D_{45}}$ & $\mathbf{D_{46}}$ & $\mathbf{D_{47}}$
		\\ \midrule
Arcene & $0.123 \; [\mathbf{0.121}]$ & $\mathbf{0.142} \; [0.146]$ & $3.199 \; [\mathbf{1.188}]$ & $\mathbf{0.179} \; [0.241]$ & $\mathbf{0.188} \; [0.224]$ & $\mathbf{0.189} \; [0.227]$ & $\mathbf{0.186} \; [0.291]$
\\
BASEHOCK & $\mathbf{6.699} \; [8.493]$ & $\mathbf{6.240} \; [8.852]$ & $110.273 \; [\mathbf{98.529}]$ & $\mathbf{10.416} \; [16.610]$ & $\mathbf{10.164} \; [15.410]$ & $\mathbf{9.911} \; [15.214]$ & $\mathbf{9.969} \; [18.831]$
\\
Caltech101 & $\mathbf{42.511} \; [79.521]$ & $\mathbf{46.142} \; [73.759]$ & $\mathbf{313.213} \; [346.309]$ & $\mathbf{72.161} \; [140.998]$ & $\mathbf{63.827} \; [129.963]$ & $\mathbf{64.586} \; [131.790]$ & $\mathbf{63.353} \; [134.950]$
\\
COIL20 & $\mathbf{1.645} \; [3.004]$ & $\mathbf{1.443} \; [2.532]$ & $18.247 \; [\mathbf{11.981}]$ & $\mathbf{2.633} \; [5.060]$ & $\mathbf{2.616} \; [4.689]$ & $\mathbf{2.725} \; [4.856]$ & $\mathbf{2.702} \; [5.053]$
\\
Isolet & $\mathbf{1.574} \; [3.124]$ & $\mathbf{1.474} \; [2.719]$ & $13.555 \; [\mathbf{11.577}]$ & $\mathbf{1.948} \; [3.856]$ & $\mathbf{1.962} \; [3.530]$ & $\mathbf{2.701} \; [4.941]$ & $\mathbf{1.894} \; [3.610]$
\\
Lung & $\mathbf{0.070} \; [0.078]$ & $\mathbf{0.066} \; [0.076]$ & $1.137 \; [\mathbf{0.789}]$ & $\mathbf{0.100} \; [0.142]$ & $\mathbf{0.099} \; [0.132]$ & $\mathbf{0.099} \; [0.132]$ & $\mathbf{0.099} \; [0.157]$
\\
Madelon & $\mathbf{3.902} \; [7.746]$ & $\mathbf{4.016} \; [7.118]$ & $\mathbf{26.026} \; [38.706]$ & $\mathbf{6.635} \; [13.229]$ & $\mathbf{6.599} \; [12.313]$ & $\mathbf{6.647} \; [12.172]$ & $\mathbf{6.537} \; [12.581]$
\\
MPEG7 & $\mathbf{1.299} \; [2.233]$ & $\mathbf{1.433} \; [2.309]$ & $12.102 \; [\mathbf{11.277}]$ & $\mathbf{2.239} \; [4.030]$ & $\mathbf{2.208} \; [3.700]$ & $\mathbf{2.229} \; [3.722]$ & $\mathbf{2.187} \; [3.929]$
\\
MPEG7-BAS & $\mathbf{0.873} \; [1.876]$ & $\mathbf{0.992} \; [1.788]$ & $\mathbf{3.530} \; [6.102]$ & $\mathbf{1.524} \; [3.158]$ & $\mathbf{1.494} \; [2.896]$ & $\mathbf{1.505} \; [2.876]$ & $\mathbf{1.490} \; [2.889]$
\\
MPEG7-Fourier & $\mathbf{0.693} \; [1.415]$ & $\mathbf{1.007} \; [1.728]$ & $\mathbf{2.211} \; [3.397]$ & $\mathbf{1.007} \; [1.966]$ & $\mathbf{1.066} \; [1.952]$ & $\mathbf{1.153} \; [2.113]$ & $\mathbf{1.001} \; [1.814]$
\\
Mushrooms & $\mathbf{31.431} \; [66.240]$ & $\mathbf{42.497} \; [64.816]$ & $\mathbf{96.601} \; [182.248]$ & $\mathbf{58.832} \; [116.774]$ & $\mathbf{52.644} \; [109.467]$ & $\mathbf{51.781} \; [107.448]$ & $\mathbf{51.858} \; [106.174]$
\\
NTL-Commercial & $\mathbf{11.692} \; [27.409]$ & $\mathbf{9.139} \; [20.447]$ & $\mathbf{22.413} \; [62.528]$ & $\mathbf{19.867} \; [43.756]$ & $\mathbf{19.758} \; [40.793]$ & $\mathbf{20.355} \; [40.969]$ & $\mathbf{20.070} \; [39.455]$
\\
NTL-Industrial & $\mathbf{5.488} \; [11.537]$ & $\mathbf{4.612} \; [9.008]$ & $\mathbf{9.879} \; [26.141]$ & $\mathbf{8.961} \; [18.301]$ & $\mathbf{9.104} \; [17.482]$ & $\mathbf{9.245} \; [17.400]$ & $\mathbf{8.919} \; [16.447]$
\\
ORL & $\mathbf{0.117} \; [0.180]$ & $\mathbf{0.133} \; [0.198]$ & $1.140 \; [\mathbf{0.882}]$ & $\mathbf{0.171} \; [0.296]$ & $\mathbf{0.176} \; [0.291]$ & $\mathbf{0.184} \; [0.301]$ & $\mathbf{0.185} \; [0.318]$
\\
PCMAC & $\mathbf{4.775} \; [6.897]$ & $\mathbf{4.594} \; [6.901]$ & $70.201 \; [\mathbf{69.133}]$ & $\mathbf{7.647} \; [13.079]$ & $\mathbf{7.396} \; [12.182]$ & $\mathbf{7.348} \; [12.059]$ & $\mathbf{7.358} \; [14.326]$
\\
Phishing & $\mathbf{60.990} \; [130.984]$ & $\mathbf{56.315} \; [84.135]$ & $\mathbf{155.342} \; [340.156]$ & $\mathbf{113.497} \; [225.509]$ & $\mathbf{101.446} \; [210.205]$ & $\mathbf{99.828} \; [205.620]$ & $\mathbf{100.060} \; [197.943]$
\\
Segment & $\mathbf{2.669} \; [5.898]$ & $\mathbf{2.355} \; [4.530]$ & $\mathbf{5.748} \; [13.794]$ & $\mathbf{4.638} \; [9.449]$ & $\mathbf{4.693} \; [8.893]$ & $\mathbf{4.761} \; [8.804]$ & $\mathbf{4.718} \; [8.472]$
\\
Semeion & $\mathbf{1.416} \; [2.914]$ & $\mathbf{1.301} \; [2.332]$ & $\mathbf{7.064} \; [9.041]$ & $\mathbf{2.451} \; [4.793]$ & $\mathbf{2.483} \; [4.533]$ & $\mathbf{2.495} \; [4.593]$ & $\mathbf{2.447} \; [4.511]$
\\
Sonar & $\mathbf{0.039} \; [0.050]$ & $\mathbf{0.045} \; [0.053]$ & $\mathbf{0.075} \; [0.115]$ & $\mathbf{0.049} \; [0.076]$ & $\mathbf{0.049} \; [0.071]$ & $\mathbf{0.052} \; [0.075]$ & $\mathbf{0.050} \; [0.071]$
\\
Spambase & $\mathbf{11.014} \; [23.017]$ & $\mathbf{9.745} \; [19.027]$ & $\mathbf{24.913} \; [57.549]$ & $\mathbf{17.983} \; [38.419]$ & $\mathbf{18.408} \; [35.998]$ & $\mathbf{17.724} \; [35.780]$ & $\mathbf{17.596} \; [34.604]$
\\
Vehicle & $\mathbf{0.334} \; [0.685]$ & $\mathbf{0.331} \; [0.601]$ & $\mathbf{0.682} \; [1.596]$ & $\mathbf{0.560} \; [1.086]$ & $\mathbf{0.564} \; [1.016]$ & $\mathbf{0.580} \; [1.051]$ & $\mathbf{0.566} \; [1.008]$
\\
Wine & $\mathbf{0.030} \; [0.039]$ & $\mathbf{0.029} \; [0.031]$ & $\mathbf{0.048} \; [0.083]$ & $\mathbf{0.042} \; [0.060]$ & $\mathbf{0.042} \; [0.057]$ & $\mathbf{0.042} \; [0.056]$ & $\mathbf{0.042} \; [0.055]$
		\\ \bottomrule
	\end{tabular}}
\end{sidewaystable}

The best OPF + Numba performance was achieved in the most time-consuming dataset (Phishing), where it could perform the training procedure in roughly $60\%$ less time than the standard OPF classifier. Moreover, we can highlight that the best OPF + Numba performances were achieved in the most extensive datasets, such as Caltech101, Mushrooms, NTL-Commercial, and Spambase, achieving up to $50\%$ less training time than the Numba-less version.

Regarding the mid-sized datasets, such as BASEHOCK, COIL20, Isolet, Madelon, MPEG7, MPEG7-BAS, MPEG7-Fourier, NTL-Industrial, PCMAC, Segment, and Semeion, we can also remark that OPF + Numba achieved a better performance than the traditional OPF, achieving up to $30-40\%$ less training time. On the other hand, when evaluating the small-sized datasets, such as Arcene, Lung, ORL, Sonar, Vehicle, and Wine, there is a slight performance increase when comparing Numba and Numba-less classifiers. Additionally, in such datasets, it is possible to observe that the Numba-less classifier achieved a shorter training time on some occasions. Such behavior is expected in faster executions due to the overhead that the Numba package provides in the code, where it performs an additional translation of Python and Numpy to faster machine code.
\section{Conclusions}
\label{s.conclusion}

This work introduces a straightforward speed up when using the Python-based Optimum-Path Forest framework, known as OPFython. Based on the single addition of the Numba package, OPFython + Numba could achieve an increased performance rate, diminishing both training and prediction times.

The OPFython + Numba framework has been assessed using a complete experimental setup, where each available distance measure has been extensively evaluated over $22$ datasets. Each experiment has been exhaustively appraised through $25$ runnings and statistically analyzed by the Wilcoxon signed-rank test. The experimental results depicted a gain of $50\%$ in performance when using OPFython + Numba compared to the Numba-less version. Although prediction times were not reported for the sake of space, they also benefitted from such a gain of performance.

We intend to keep fostering OPFython package by including new OPF-based classifiers and improving its base structures regarding future works. Furthermore, we aim to continuously enhance its distance calculation, attempting to provide more competitive performance than the C-based package.

% Acknowledgments
\section*{Acknowledgments}
The authors are grateful to S\~ao Paulo Research Foundation (FAPESP) grants \#2013/07375-0, \#2014/12236-1, \#2019/07665-4, \#2019/02205-5, and \#2020/12101-0, and to the Brazilian National Council for Research and Development (CNPq) \#307066/2017-7 and \#427968/2018-6.

% Bibliography
\bibliographystyle{unsrt}  
\bibliography{paper}

\end{document}